\pdfoutput=1
\documentclass{ecai}

\usepackage{microtype}
\usepackage{hyperref}
\usepackage{url}
\usepackage{booktabs}

\usepackage{times}
\usepackage{latexsym}

\usepackage[T1]{fontenc}

\usepackage[utf8]{inputenc}

\usepackage{enumerate}
\usepackage{graphicx}
\usepackage[skip=0pt]{caption}
\usepackage{subcaption}
\usepackage{graphics, amsmath}
\usepackage{amssymb}
\usepackage{multirow}

\usepackage{tabularx}
\usepackage{pifont}
\usepackage{colortbl}
\usepackage{xspace}
\usepackage{xcolor}
\usepackage[most]{tcolorbox}
\usepackage{makecell}
\usepackage{adjustbox}
\usepackage[inline]{enumitem}

\usepackage{arydshln}
\usepackage{acronym}

\usepackage{wrapfig}

\usepackage{floatrow}
\newfloatcommand{capbtabbox}{table}[][\FBwidth]

\usepackage{blindtext}
\newtcolorbox[list inside=prompt,auto counter,number within=section]{prompt}[1][]{
    colbacktitle=black!60,
    coltitle=white,
    fontupper=\footnotesize,
    boxsep=5pt,
    left=0pt,
    right=0pt,
    top=0pt,
    bottom=0pt,
    boxrule=1pt,
    #1,
}

\newcommand{\cmark}{\ding{51}}
\newcommand{\xmark}{\ding{55}}

\newcommand{\nsampletotal}{$4,909$\xspace}

\newcommand{\nsampletrain}{$3,871$\xspace}
\newcommand{\nsamplevalid}{$430$\xspace}
\newcommand{\nsampletest}{$608$\xspace}

\newcommand{\annollm}{LLM$_\text{Anno}$\xspace}
\newcommand{\rotowire}{\textsc{RotoWire}\xspace}
\newcommand{\qtsumm}{\textsc{QTSumm}\xspace}
\newcommand{\ours}{\textsc{QFMTS}\xspace}

\newcommand{\tapex}{TAPEX\xspace}
\newcommand{\omnitab}{OmniTab\xspace}

\newcommand{\multitabqa}{MultiTab\xspace}
\newcommand{\bartbasesum}{BART-base-FT\xspace}

\newcommand{\bartlargesum}{BART-large-FT\xspace}
\newcommand{\tapexlargesum}{TAPEX-FT\xspace}
\newcommand{\omnitablargesum}{OmniTab-FT\xspace}
\newcommand{\multitabqasum}{MultiTab-FT\xspace}

\newcommand{\llamasum}{Llama-2-FT\xspace}

\newcommand{\directsumm}{DirectSumm\xspace}
\newcommand{\reasonsumm}{Reason-then-Summ\xspace}

\begin{document}

\begin{frontmatter}

\title{\ours: Generating Query-Focused Summaries over Multi-Table Inputs}

\author[A]{\fnms{Weijia}~\snm{Zhang}\thanks{Email: w.zhang2@uva.nl.}\footnote{Equal contribution.}}
\author[A]{\fnms{Vaishali}~\snm{Pal}\footnotemark}
\author[A]{\fnms{Jia-Hong}~\snm{Huang}} 
\author[A]{\fnms{Evangelos}~\snm{Kanoulas}} 
\author[A]{\fnms{Maarten}~\snm{de Rijke}} 

\address[A]{Information Retrieval Lab (IRLab), University of Amsterdam}

\begin{abstract}
Table summarization is a crucial task aimed at condensing information from tabular data into concise and comprehensible textual summaries. However, existing approaches often fall short of adequately meeting users' information and quality requirements and tend to overlook the complexities of real-world queries. In this paper, we propose a novel method to address these limitations by introducing query-focused multi-table summarization. Our approach, which comprises a table serialization module, a summarization controller, and a large language model (LLM), utilizes textual queries and multiple tables to generate query-dependent table summaries tailored to users' information needs. To facilitate research in this area, we present a comprehensive dataset specifically tailored for this task, consisting of \nsampletotal query-summary pairs, each associated with multiple tables. Through extensive experiments using our curated dataset, we demonstrate the effectiveness of our proposed method compared to baseline approaches. Our findings offer insights into the challenges of complex table reasoning for precise summarization, contributing to the advancement of research in query-focused multi-table summarization.

\end{abstract}

\end{frontmatter}

\section{Introduction}
\label{sec:intro}

\begin{figure}[t!]
 \centering
    \includegraphics[width=0.95\textwidth]{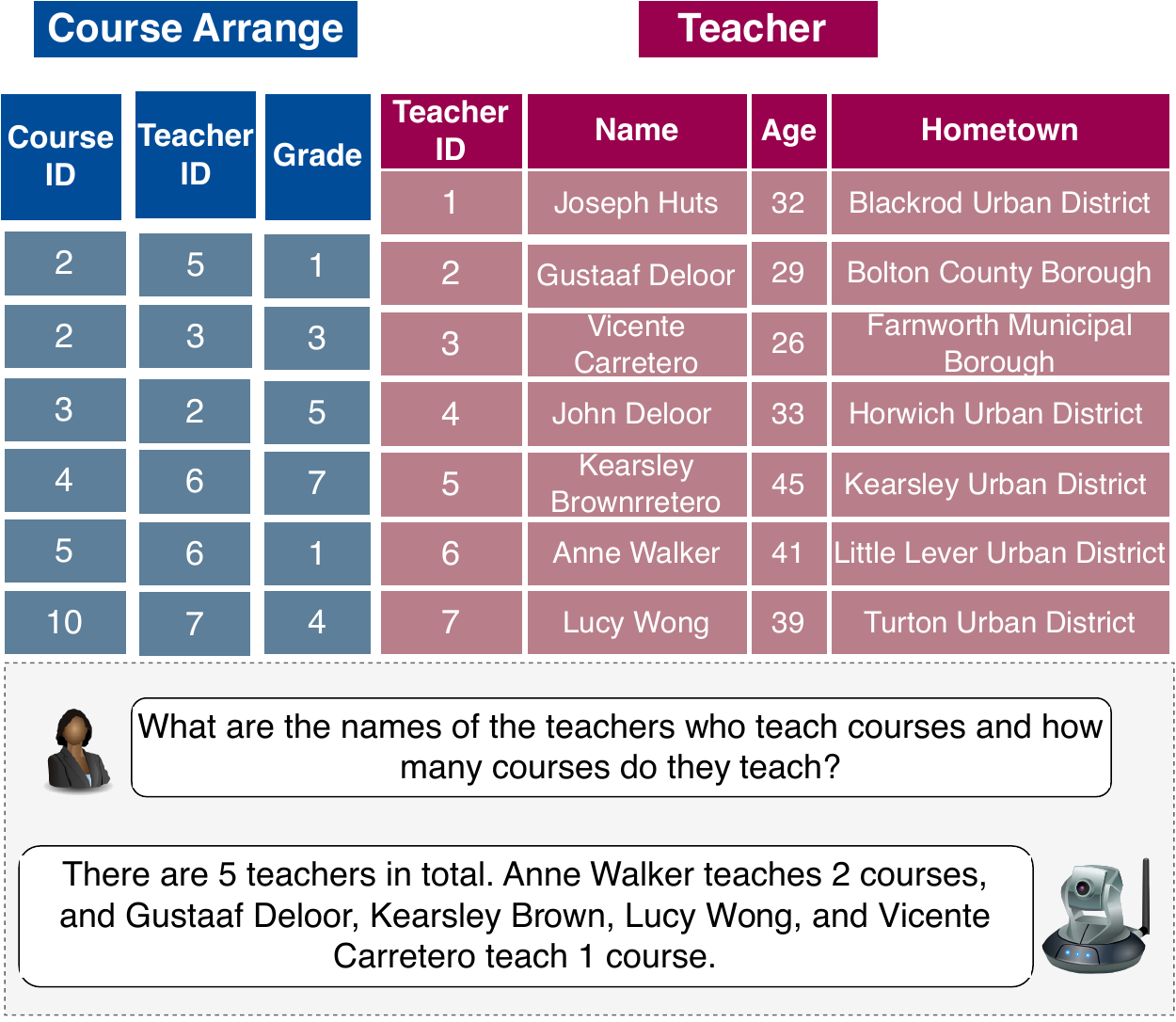}
 \caption{An example of query-focused multi-table summarization. Summarization models should combine the information from the two tables to produce a summary tailored to the query.}
 \label{fig:main_diagram}
\end{figure}

Table summarization involves the distillation of information from tabular data into a succinct format, typically a clear and human-readable description or textual table summary. This process aims to capture the key insights or trends encapsulated within the table, allowing for easier comprehension and interpretation by humans \cite{lebret-etal-2016-neural,chen-etal-2020-logical,suadaa-etal-2021-towards,liu-etal-2022-plog,liu-etal-2022-long}. Traditional methods of table summarization take a single table as input and produce a fixed textual summary \cite{jain-etal-2018-mixed,zhang2020summarize}. 
However, the fixed nature of table summaries generated by these traditional approaches often falls short of meeting users' information and quality requirements adequately. The quality of a summary plays a pivotal role in its utility. For instance, in business contexts, table summaries often play a critical role in shaping future business strategies, with their quality directly influencing the judgments of decision-makers. Poor-quality summaries may fail to capture essential aspects, leading to a misrepresentation of the data and conveying an inaccurate perspective. 

To enhance the effectiveness of conventional table summarization, \citet{zhao-etal-2023-qtsumm} proposes to leverage textual queries as a starting point to generate query-dependent table summaries aligned with users' information needs. Their approach accepts a single table and a user-specified textual query as inputs, generating a description or statement tailored to the query's focus as output. 
However, in \citet{zhao-etal-2023-qtsumm}, the authors presume that fulfilling the information requirements of a given query depends solely on data from a single table. This assumption overlooks the complexity of real-world scenarios, which frequently demand information from multiple data sources. Consequently, the intricate nature of real-world queries highlights the necessity for integrating information from multiple tables to address them effectively.

Let's consider a practical scenario depicted in \autoref{fig:main_diagram} to understand how humans address complex real-world queries.
For instance, the query ``\textit{What are the names of the teachers who teach courses and how many courses do they teach?}'' entails two distinct information requirements – teachers' names and their course teaching details. While the \texttt{Teacher} table in  \autoref{fig:main_diagram} provides the teachers' names, relying solely on this table is insufficient to generate a complete query-focused table summary. To completely address the query, we require data from another table, such as the \texttt{Course Arrange} table in \autoref{fig:main_diagram}, which contains information about course arrangements for teachers listed in the \texttt{Teacher} table. By performing table \textit{join} (i.e., multi-table reasoning) and \textit{count} (i.e., arithmetic reasoning) operations, we can determine the number of courses taught by each teacher. Thus, addressing such a common real-world complex query involves employing multi-table reasoning and arithmetic reasoning. This practical and challenging scenario remains largely unexplored, underscoring the necessity for further exploration and advancement in this domain.

\begin{figure}[t]
 \centering
    \includegraphics[width=\textwidth]{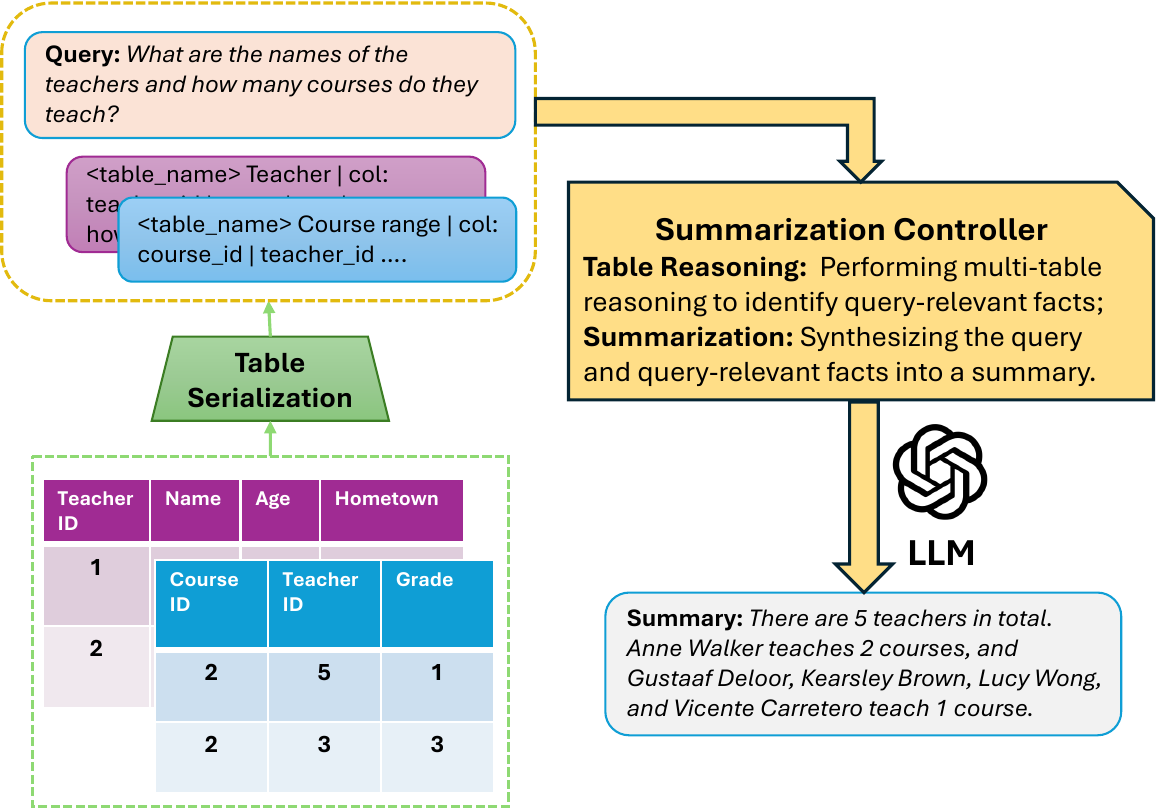}
 \caption{Overview of the proposed approach. Our approach integrates a table serialization module with a summarization controller. Initially, a set of tables is transformed into textual representations through the table serialization module. Subsequently, the summarization controller directs an LLM to perform table reasoning and produce a summary specifically tailored to a given query.}
 \label{fig:framework}
\end{figure}

Motivated by the above observation of how humans handle such a scenario and aiming to bridge this research gap, we propose a new method designed to tackle the aforementioned practical and challenging situation: query-focused multi-table summarization. Our proposed approach involves taking multiple tables and a user-defined query as inputs, aiming to generate a query-relevant textual summary based on the inputs. The proposed approach primarily comprises a table serialization module, a summarization controller, and a large language model (LLM). 
The flowchart illustrating our proposed method is provided in \autoref{fig:framework} to aid comprehension.
Additionally, in light of the absence of an existing dataset for our introduced task, and to validate the effectiveness of our proposed query-focused multi-table summarization method, we create a query-focused multi-table summarization (\ours) dataset. This dataset comprises \nsampletotal query-summary pairs, each pair associated with multiple tables. For further details about our proposed dataset, please refer to Section \ref{sec:dataset}.

In our comprehensive experiments, we assess the effectiveness of the proposed query-focused multi-table summarization method using our devised \ours dataset. The experimental findings demonstrate the superiority of our approach over baseline methods, shedding light on the challenges encountered by existing models in performing complex table reasoning to produce precise table summaries. 
To the best of our knowledge, we are the first to address the task of query-focused multi-table summarization.

\vspace{+3pt}
\noindent Our primary contributions can be summarized as follows:

\begin{itemize}[leftmargin=*]
    \item \textbf{Introduction of Query-Focused Multi-Table Summarization:} We introduce a novel method tailored for query-focused multi-table summarization, aiming to overcome limitations present in current approaches that predominantly target single-table summarization. Our proposed method comprises a table serialization module, a summarization controller, and a large language model (LLM), offering a structured framework to tackle the complexities inherent in the task of query-focused multi-table summarization.

    \item \textbf{Development of a Comprehensive Dataset:} We develop a \ours dataset specifically tailored for the query-focused multi-table summarization task. The \ours dataset consists of $4,909$ query-summary pairs, each intricately linked with multiple tables. It will serve as a valuable resource for researchers in this field, facilitating the exploration and validation of proposed methods in the future.

    \item \textbf{Extensive Experiments:} We conduct comprehensive experimental validation of our proposed query-focused multi-table summarization method using the introduced \ours dataset. Our experimental results demonstrate its effectiveness over baseline methods and provide insights into the challenges encountered in complex table reasoning for precise summarization.
\end{itemize}

\section{Related Work}
\label{related}

In this section, we briefly review recent developments in the related areas of table summarization and query-focused summarization. We also highlight our unique contributions in contrast to these existing studies.

\paragraph{Table Summarization}

Table summarization involves generating a concise and informative summary from a given table. To address this task, prior research~\cite{zhang2020summarize,jain-etal-2018-mixed,liu-etal-2022-long} has primarily focused on summarizing the entire table without explicitly addressing users' specific information needs. However, in real-world applications, users frequently seek targeted information from table segments, underscoring the importance of generating personalized table summaries. Although \citet{zhao-etal-2023-qtsumm} introduced the initial human-annotated dataset for query-focused table summarization, their study restricted to single-table scenarios, lacking consideration of multi-table reasoning such as operations involving table \textit{join} and \textit{union}. In contrast. our work presents two primary distinctions: first, we leverage LLMs to assist in the data annotation; second, our proposed method addresses complex queries that necessitate the integration of information across multiple table contexts.

\paragraph{Query-Focused Summarization}

Query-focused summarization is designed to generate textual summaries based on a specific query and a collection of relevant contexts. The contexts can be either textual~\cite{dang-2006-duc,zhang2021scaling,zhang2023kiqfs,zhang2024comparative,zhang2024towards,zhang2024beyond}, or visual data~\cite{huang2020query,huang2021gpt2mvs,huang2021deepopht,huang2022causal,huang2023causalainer,huang2023conditional,huang2023improving,huang2023query,huang2024novel,huang2024multi,zhu2024enhancing}. 
Traditional studies~\cite{huang2017vqabq,huang2018robustness,huang2019novel,huang2019assessing,xu-lapata-2020-coarse,xu-lapata-2021-generating} have faced challenges due to the scarcity of large-scale datasets, often resorting to distant supervision signals from adjacent fields such as generic summarization to enhance summarization performance. Additionally, recent efforts have been directed towards creating synthetic large-scale datasets~\cite{kulkarni2020aquamuse,laskar2023cqsumdp,liu2024querysum,huang2024optimizing}. Despite these advancements, the application of query-focused summarization to tabular data remains relatively unexplored~\cite{zhao-etal-2023-qtsumm}, especially in scenarios where queries span multiple tables. This study seeks to bridge this gap by exploring the efficacy of query-focused summarization in multi-table contexts.

\section{Methodology}
\label{sec:method}

This section formulates the task of query-focused multi-table summarization and details our proposed approach. As illustrated in \autoref{fig:framework}, our approach comprises two primary components: a table serialization module and a summarization controller.

\paragraph{Task Formulation}
\label{subsec:task}

Query-focused multi-table summarization, denoted as \ours, involves generating a coherent and informative summary aimed at addressing a user query across multiple tables. Specifically, given a natural language query $q$ and a set of input tables $\mathcal{T}={t_1, \ldots, t_k}$, a query-focused multi-table summarization model systematically engages in table-based reasoning across the contents of $\mathcal{T}$ related to $q$, aiming to produce a textual summary $s$ that effectively resolves the user query while maintaining factual accuracy and comprehensiveness.

\paragraph{Table Serialization}
\label{subsec:table_format}

Given that our approach is based on LLMs which process only textual data, it necessitates a table serialization to transform input tabular data into a textual format suitable for processing.
In this work, we utilize a technique known as table linearization, which is widely used in table-to-text generation tasks~\cite{liu2022tapex,liu-etal-2022-plog,nan-etal-2022-fetaqa,pal-etal-2022-parameter}. This technique transforms a table into a textual, sequential format using designated sentinel words. Specifically, a table identified by its name, $name$, consisting of $m$ rows and $n$ columns, is linearized as follows:
\begin{align*}
& \textbf{<table\_name>:}\ name \qquad \ \textbf{col:} \ h_1 \mid \ldots \mid h_n \\ 
& \textbf{row\ 1:}\ c_{1,1} \mid \ldots \mid c_{1,n} \ldots \quad \textbf{row\ m:}\ c_{m,1} \mid \ldots \mid r_{m,n}.
\end{align*}
\noindent
where $h_j$ and $c_{i,j}$ stand for the $j$-th column header and $i$-th row and $j$-th column cell, respectively.

\begin{figure}[t]
 \centering
    \includegraphics[width=\textwidth]{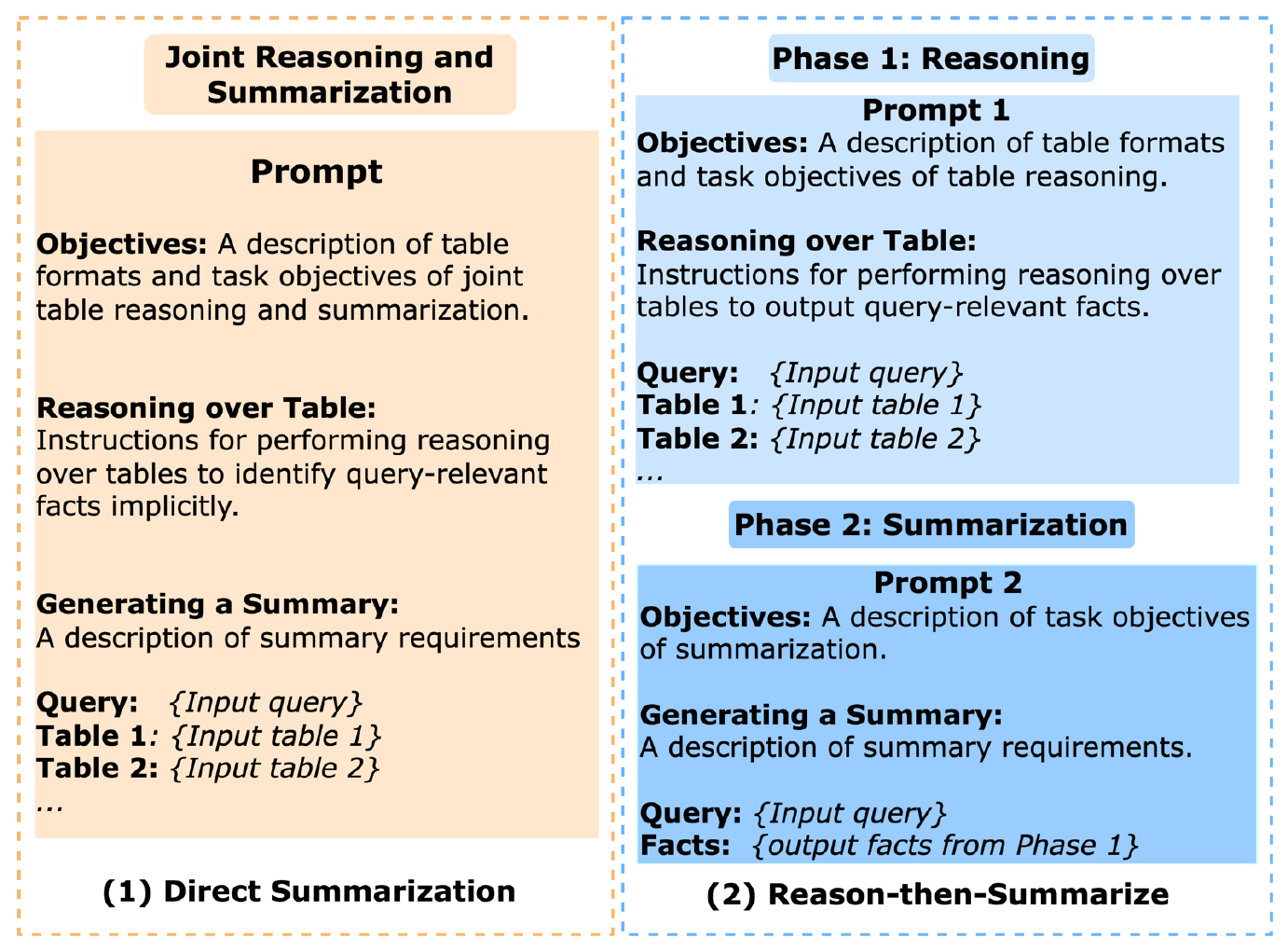}
 \caption{Two distinct approaches of the summarization controller: (1) Direct Summarization (2) Reason-then-Summarize. The former tackles the proposed task in an end-to-end manner. In contrast, the latter tackles the task across two independent phases.}
 \label{fig:main_method}
\end{figure}

\subsection{Summarization Controller}
\label{subsec:summ}

Given a query and the corresponding linearized tables, our proposed summarization controller efficiently generates a comprehensive summary. This is achieved by integrating an LLM, such as GPT-3.5~\cite{wang2022rlhf}, with a carefully designed prompting-based method. In this work, we explore two distinct methods: \textit{direct summarization} and \textit{reason-then-summarize}, which are depicted in \autoref{fig:main_method}. The complete prompts for the proposed methods are demonstrated in Appendix~\ref{app_sec:propmt}.

\paragraph{Direct Summarization}

Direct summarization (\directsumm) enables the LLM to jointly perform table reasoning and summarization in an end-to-end manner. We first present the linearized structure of table formats as outlined in Section \ref{subsec:table_format}, aiming to improve the LLM's understanding of tabular data. Given the query and linearized tables, we then prompt the LLM to perform table reasoning across multiple tables to identify query-relevant facts implicitly. i.e., these facts are not explicitly generated by the LLM. It is worth noting that the facts include numerals and entities that address the information needs of the query. Furthermore, we adopt chain-of-thought (CoT) style prompting methods~\cite{wei2022cot,kojima2022zerocot} using the directive \textit{``Let's think step by step''}. This enhances the LLM's reasoning ability. Lastly, we instruct the LLM to synthesize the query with the identified query-relevant facts into a comprehensive summary, while adhering to constraints such as summary length.

\paragraph{Reason-then-Summarize}

Inspired by the established \textit{retrieve-then-generate} paradigm~\cite{lewis2020rag,izacard-grave-2021-leveraging}, we introduce a novel approach termed \textit{reason-then-summarize} (\reasonsumm). This approach tackles the proposed task in sequential phases. The first phase exclusively focuses on enabling the LLM to perform table reasoning across multiple tables based on the query and linearized tables. As a result, the LLM identifies and extracts query-relevant facts from the tables. In the second phase, we prompt the LLM to synthesize a comprehensive summary based on the query and the previously extracted query-relevant facts.

\begin{table*}[t!]
\caption{Comparisons between our dataset and existing table summarization datasets. Our dataset is the only one tailored for query-focused multi-table summarization, supporting both numeric and multi-table reasoning.} 
\label{tab:datasets_comparison}
\centering
\begin{adjustbox}{max width=0.985\textwidth} {
\begin{tabular}{lccccc}
\toprule
 \multirow{3}{*}{\textbf{Dataset}} & \multicolumn{3}{c}{\textbf{Statistics}} & \multicolumn{2}{c}{\textbf{Reasoning}} \\
  \cmidrule(r){2-4} \cmidrule{5-6}
  &  \# Example & \makecell{\# Table per Example} &  \makecell{\# Words in Summary} & Numeric & \makecell{Multi-Table}  \\
 \midrule
\rotowire~\citep{lebret-etal-2016-neural} & \phantom{0}4,953 & 1.0\phantom{0} & 337.1 & \xmark & \xmark \\
SciGen~\citep{moosavi2021scigen} & \phantom{0}1,338 & 1.0\phantom{0} & 116.0 & \cmark & \xmark \\
NumericNLG~\citep{suadaa-etal-2021-towards} & \phantom{0}1,355 & 1.0\phantom{0} & \phantom{0}94.2 & \cmark & \xmark \\
\qtsumm~\citep{zhao-etal-2023-qtsumm} & \phantom{0}7,111 & 1.0\phantom{0} & \phantom{0}68.0  & \cmark & \xmark \\
\midrule
\ours & \phantom{0} \nsampletotal & 1.83 & \phantom{0}58.5 & \cmark & \cmark \\
\bottomrule
\end{tabular}}
\end{adjustbox}
\end{table*}

\section{Dataset Construction}
\label{sec:dataset}

In order to validate the effectiveness of our proposed method, we create a novel dataset specifically tailored for this query-focused multi-table summarization task. This section details the dataset construction process, including data annotation and quality verification.

\paragraph{Source Data} 
\label{subsec:source}

We build our dataset on top of the Spider dataset~\cite{yu-etal-2018-spider}. 
The Spider dataset, originally designed for semantic parsing and text-to-SQL tasks, comprises $10,181$ natural language queries. These queries are paired with complex SQL queries and one or more tables from various relational databases.
In this work, we construct input queries and tables from original textual queries and their corresponding tables to develop our dataset. We utilize SQL queries for data annotation, as detailed in the following subsection.
Additionally, our analysis shows that over $50\%$ of queries in the original dataset have only a single table. This high proportion of single-table inputs raises concerns about the dataset's efficacy to provide sufficient challenges for multi-table scenarios. To address this, we selectively down-sampled these single-table examples to allow multi-table examples to predominate in the dataset. Since the test set from the original dataset is not publicly available, we randomly allocated $10\%$ of the original training set to serve as our validation set, with the remaining $90\%$ forming the new training set. The original validation set was repurposed as our test set. This results in \nsampletrain\ training, \nsamplevalid\ validation, and \nsampletest\ test examples, forming the basis for our dataset.

\subsection{Data Annotation}
\label{subsec:annotation}

\paragraph{LLMs as Data Annotators}

The objective of data annotation is to produce high-quality, comprehensive, and accurate summaries tailored to the associated input queries.
Prior research in the field of query-focused summarization from single tables, such as \qtsumm~\cite{zhao-etal-2023-qtsumm}, has predominantly depended on human experts to annotate summaries based on an input query and an input table. This reliance is primarily due to the complex table reasoning, making summary annotation a challenging task.
However, manual annotations are not only time-consuming but also costly. Recent studies~\cite{ding-etal-2023-gpt,zhao-etal-2023-investigating,he2023annollm,yu2023llmattr} have revealed that LLMs can match the annotation quality of crowd-sourced workers while being significantly more cost-effective and efficient. Motivated by these findings, we have employed LLMs as data annotators for our summary annotation, namely \annollm. Although \annollm does not yet match the table reasoning capabilities of human experts, we have designed a simplified table-to-text annotation task avoiding complex table reasoning.
Specifically, for each textual query in our dataset, we follow \citet{pal-etal-2023-multitabqa} to first extract the \textit{output table} by executing the corresponding SQL query over the associated rational database. It is worth noting that the execution table contains query-relevant facts/entities required to construct a summary. Subsequently, rather than relying on input tables, we use the execution table as the basis for our annotation. 
We also have identified that relying solely on the output table frequently results in summaries that lack essential contextual information. Our observations indicate that this missing context can effectively be retrieved directly from the input query.
For instance, consider the example output table below:
\begin{center}
\begin{minipage}[c]{0.6\columnwidth}
\small
\centering
\begin{tabular}{|c|c|}
\hline
semester\_name & semester\_id \\
\hline
summer 2010 & 2 \\
\hline 
\end{tabular}
\end{minipage} 
\end{center}
It lacks contextual information to confirm that the \texttt{Summer 2010} semester has the \textit{most registered students} in response to the corresponding query \textit{``What is the semester in which most students registered? Show both the name and the ID.''}.
To address this, we have incorporated the given query as a supplementary input for summary annotation.
This strategy enhances the overall comprehensiveness and relevance of the annotated summaries. 
In this work, we employ \texttt{gpt-3.5-turbo-0613} as \annollm via the public OpenAI API.\footnote{\url{https://platform.openai.com}}. 

\begin{prompt}[title={Prompt \thetcbcounter: Summary Annotation}, label=prompt:annotation]
\textbf{Instruction:} A comprehensive annotation guideline. \\
\\
\textbf{Demonstrations:} \\
Few-shot human-written demonstrations. \\
\\ 
\textbf{Query:} \{\textit{Input query}\} \\
\textbf{Table:} \{\textit{Linearized execution table}\}
\end{prompt}

\paragraph{Instruction Design}
\label{para:sum_inst}

The effectiveness of the instruction prompt is crucial in determining the quality of annotated summaries. To this end, we carefully design the instruction to ensure the summary quality, in which the structure of the instruction is shown in Prompt~\autoref{prompt:annotation}. The prompt comprises three components: a comprehensive annotation guideline, few-shot demonstrations, and input data. 
The annotation guideline outlines the expected discourse structure and the summary's length requirements. We have found that a more precise guideline significantly improves generation quality. To provide further clarity to \annollm, we manually write summaries for a few examples as few-shot demonstrations (we use $5$-shot in our experiments). Lastly, we include both the input query and the execution table directly in the prompt, specifically requesting \annollm to write a summary based on these inputs. Similarly, we leverage table serialization as described in Section \ref{subsec:table_format} to obtain the linearized execution table.

\subsection{Dataset Analysis}

Our dataset comprises a total of \nsampletotal examples, segmented into \nsampletrain training examples, \nsamplevalid validation examples, and \nsampletest test examples. 
The dataset is composed of $32.8\%$ single-table examples, $52.6\%$ double-table examples, and $14.6\%$ examples that incorporate three or more tables. Notably, more than $67\%$ of the examples include at least two tables, highlighting the dataset's efficacy in facilitating research in multi-table scenarios.
We present a comparative analysis of our dataset against existing table-to-text generation datasets in \autoref{tab:datasets_comparison}. Our dataset averages approximately two input tables per example, in contrast to the prevailing datasets which predominantly focus on single-table scenarios. The summaries from our dataset are sufficiently informative, with an average length of $58.5$ words, aligning closely with the norms of existing datasets.
Additionally, our dataset is characterized by a rich variety of operations. It includes basic numeric operations such as \textit{sum} and \textit{average}, and extends to more complex multi-table operations like \textit{join} and \textit{union}, which are absent in the \qtsumm dataset, as it is tailored exclusively towards single-table contexts.

\subsection{Quality Verification and Control}
\label{subsec:quality_eval}

To assess the quality of annotated summaries effectively, we develop a comprehensive evaluation encompassing both automated and manual evaluations. We define three primary desiderata for quality verification:
\begin{itemize}[leftmargin=*,nosep]
    \item \textbf{Faithfulness:} Each statement within the summary must be factually consistent with the facts presented in the execution table.
    \item \textbf{Completeness:} The summary should address all information needs in the user query, representing all facts from the execution table.
    \item \textbf{Fluency:} The summary needs to be articulate, clear, and easily understandable for human readers.
\end{itemize}
\noindent %
In our experiments, we utilize standard sequence similarity metrics to evaluate completeness. Given the lack of definitive metrics for faithfulness and fluency, we rely on human evaluations to assess these aspects. The results are presented in \autoref{tab:quality_eval}.

\begin{table}[t!]
     \caption{Quality evaluation of our dataset. $^*$represents that the completeness is quantified using average ROUGE-L recall scores. $^\dagger$represents that we randomly sample 100 examples from the training, validation, and test set to measure faithfulness (0-1) and fluency (1-5), respectively.}
    \label{tab:quality_eval}
    \centering
    \begin{adjustbox}{max width=0.95\textwidth} {
         \begin{tabular}{l *{3}{c}}
        \toprule
        \multirow{2}{*}{\textbf{Split}}  & \textbf{Automated Evaluation} & \multicolumn{2}{c}{\textbf{Human Evaluation$^\dagger$}} \\
        \cmidrule(r){2-2} \cmidrule{3-4}
        & Completeness$^*$ (\%) & Faithfulness  & Fluency \\
        \hline
        Training   & 91.45  & 0.98 & 4.73 \\
        Validation   & 91.75  & 0.98 & 4.81 \\
        Test    & 90.75  & 0.96 & 4.68 \\
        Total   & 91.48  & 0.97 & 4.74 \\
        \bottomrule
        \end{tabular}
    } 
    \end{adjustbox}
\end{table}

\paragraph{Automated Evaluation}
\label{para:auto_eval}

We initially assess the completeness of the annotated summary solely based on the execution table. However, as delineated in Subsection \ref{subsec:annotation}, the execution table, by itself, proves inadequate for a comprehensive evaluation of summary completeness due to its limited contextual information. Consequently, we extend our evaluation by incorporating the query with the table, thereby enabling a more robust measure of completeness. Specifically, we extract facts, including numerals and entities, from the execution table. These facts are then combined with the query to construct a reference sequence. The completeness of the annotated summary is estimated against this reference sequence.
In this work, we quantify the completeness using the lexical similarity metric ROUGE-L~\cite{lin-2004-rouge}, a standard metric in table-to-text generation assessments~\cite{lin2023survey,zhao-etal-2023-qtsumm}. As our primary focus lies in assessing the presence of information from the query and the execution table within the summary, we focus on the recall scores of ROUGE-L. As shown in \autoref{tab:quality_eval}, the ROUGE-L recall scores exceed $90$, affirming that the annotated summaries proficiently include not only the facts from the corresponding execution tables but also the contextual information from the queries.

\paragraph{Human Evaluation}

To assess the faithfulness and fluency of annotated summaries, we randomly select $100$ examples from each of the training, validation, and test sets. We engage three annotators, each proficient in SQL and English, to evaluate the summaries in relation to the corresponding SQL queries, input tables, and execution tables.
Annotators assign a binary label to assess faithfulness, a common method in table-to-text generation tasks~\cite{chen2020tabfact,zhao-etal-2023-qtsumm}. Summaries that accurately represent the execution tables without any hallucinated content are labeled as $1$, while those that do not are labeled as $0$. The annotators are also provided with the query and input tables to enhance their understanding and judgment of the summaries.
Following \citet{zhao-etal-2023-qtsumm}, we measure fluency using a $5$-point Likert scale, ranging from $1$ (least fluent) to $5$ (most fluent). 
The average score from the three annotators determines the faithfulness and fluency rating for each summary.
\autoref{tab:quality_eval} presents the results: the summaries achieved an average faithfulness score of $0.97$ and a fluency score of $4.74$, indicating that over $97\%$ of the summaries are faithful to the corresponding execution tables and are deemed sufficiently fluent by the annotators.
To measure the inter-annotator agreement, we employed the Fleiss Kappa scores \cite{fleiss1971measuring}, achieving Kappa scores of $0.97$ for faithfulness and $0.80$ for fluency. These scores indicate almost perfect agreement and substantial agreement, respectively.

\paragraph{Human Post-Correction}

Despite our quality verification confirming the high quality of the dataset, we acknowledge the need for further correction of the validation and test sets to ensure their accuracy, as they play a critical role in selecting optimal model checkpoints and measuring model performance, respectively. Additionally, biases may arise from using output summaries produced by \annollm to construct these sets. This risk is particularly pronounced if the same \annollm is employed as the baseline model, potentially leading to artificially enhanced performance results. To address these concerns, we have implemented a rigorous post-correction process on the annotated summaries within both sets. This involves a detailed manual review to identify and rectify any missing information or hallucinated content in the summaries, based on the corresponding query and execution table. Furthermore, we have undertaken to rephrase each summary in a manner that more closely resembles human expression, thereby reducing potential biases. See Appendix \ref{app_subsec:correction_details} for more details about human post-correction.

\section{Experiment}
\label{sec:setup}

In this section, we outline baseline models selected for performance comparison. We then provide the implementation details of the baseline models and our proposed method. Lastly, we describe the evaluation protocols employed in our experiments.

\subsection{Baseline Models}
\label{subsec:models}

In this work, we conduct experiments to evaluate two distinct neural network architectures: encoder-decoder models and decoder-only LLMs. Given that encoder-decoder models typically have significantly fewer parameters than LLMs, we fine-tune these models using our training dataset to facilitate a fair comparison. In contrast, we utilize the LLMs as backbone models for our proposed methods without updating their parameters in the experiments. We benchmarked against the following state-of-the-art models:

    \smallskip\noindent%
    \textbf{BART}~\cite{lewis-etal-2020-bart} represents a pre-trained encoder-decoder architecture known for its efficacy in text generation tasks. We have fine-tuned two variants of BART, namely \texttt{bart-base} with 139 million parameters and \texttt{bart-large} with 406 million parameters~\cite{lewis-etal-2020-bart}, which are referred to as \bartbasesum and \bartlargesum, respectively. 

    \smallskip\noindent%
    \textbf{\tapex}~\cite{liu2022tapex} is a table-to-text generation model, trained using a large-scale synthetic dataset that includes executable SQL queries and their corresponding outputs. In our experiments, we employ the version that utilizes \texttt{bart-large} as the backbone.

    \smallskip\noindent%
    \textbf{\omnitab}~\cite{jiang-etal-2022-omnitab} is based on the same architecture as \tapex but has been additionally trained on a synthetic dataset designed for table question answering tasks. Like \tapex, our implementation leverages \texttt{bart-large} as its backbone.

    \smallskip\noindent%
    \textbf{\multitabqa}~\cite{pal-etal-2023-multitabqa} is a table-to-text generation model that has been additionally trained on a synthetic, multi-table question answering dataset. In our experiments, we utilize \texttt{bart-base} as its backbone, as the released version exclusively supports this backbone.

    \smallskip\noindent%
    \textbf{Llama-2}~\cite{touvron2023llama2} includes a set of open-source LLMs that have been pre-trained across vast datasets. We explore two adaptations of Llama-2. First, we fine-tune it on our dataset, namely \llamasum. Second, we utilize llama-2 as the backbone for our approach.

    \smallskip\noindent%
    \textbf{GPT}~\cite{wang2022rlhf,openai2023gpt4} comprises a family of LLMs developed by OpenAI, demonstrating their remarkable text generation capabilities across numerous tasks. In our experiments, we deploy versions \texttt{gpt-3.5-turbo-0613} and \texttt{gpt-4-0613}, applying these models as the backbones for our approach.

\subsection{Implementation Details}
\label{app_subsec:imple_details}

\paragraph{Fine-Tuning}

We fine-tune encoder-decoder models on the training set using the AdamW optimizer~\cite{loshchilov2019adamw}. This is conducted over $32$ epochs with a learning rate of $1e^{-4}$, batch size of $256$, and the maximum sequence length of $1024$. 
For fine-tuning Llama-2 models, we adopt the QLoRA algorithm~\cite{dettmers2023qlora} to fine-tune \texttt{llama-2-chat-7B} due to the computation restrictions. The maximum sequence length is $4096$.
Input sequences for the fine-tuned models were composed by concatenating the query with all linearized input tables. For instance, the final input sequence for an example with $k$ tables is represented as $query\ [table_1] \ldots [table_k]$, where $[table_i]$ is the linearized representation of the \textit{i}-th input table.
Model performance is evaluated by selecting the best checkpoints based on the loss from the validation set. All experiments are conducted on a single A6000 GPU. 

\paragraph{Few-Shot Prompting}

We prepend $3$-shot demonstrations into the prompts to facilitate in-context learning~\cite{brown2020gpt3} for our methods. We set the temperature, top-p, and maximum output tokens to $0.1$, $0.95$ and $400$, respectively. Due to budget constraints, we only report the results of the \reasonsumm method with the backbone of GPT-4.

\begin{table*}[ht]
\caption{Summarization performance of our approaches with various backbones and fine-tuned models on the test set of our dataset. ``FT'' stands for the fine-tuned version of the corresponding model. The best results are highlighted in \textbf{bold}. $^*$represents that the released version only supports the backbone of BART-base.}
\label{tab:main_results}
\centering
\begin{adjustbox}{max width=0.985\textwidth} {
\begin{tabular}{l *{6}{c}}
    \toprule
        \multirow{2}{*}{\textbf{Model}} & \multirow{2}{*}{\textbf{Backbone}}  & \multicolumn{3}{c}{\textbf{Text-Based Metric}}  & \multicolumn{2}{c}{\textbf{Table-Based Metric}} \\ 
        \cmidrule(lr){3-5} \cmidrule{6-7}
       & & \textbf{SacreBLEU} & \textbf{ROUGE-L} & \textbf{BERTScore} & \textbf{STR-EM} & \textbf{PARENT}  \\
    \midrule
    \textit{Fine-tuned} \\
        \quad \bartbasesum & BART-base & $39.74$ & $63.14$ & $62.38$  & $24.63$ & $13.51$ \\
        \quad \bartlargesum & BART-large & $43.40$ & $64.84$ & $66.06$  & $33.54$ & $18.66$ \\
        \quad \tapexlargesum & BART-large & $43.99$ & $65.12$ & $66.43$  & $38.78$  & $22.16$ \\
        \quad MultiTab-FT$^*$  & BART-base & $44.41$ & $65.68$ & $67.13$  & $42.70$  & $24.83$ \\
        \quad \omnitablargesum & BART-large  & $45.58$ & $67.19$ & $68.76$  & $44.60$ & $26.46$ \\
        \quad \llamasum & Llama-2-7B  &  $\textbf{54.06}$ & $\textbf{71.82}$ & $\textbf{73.66}$ & $61.71$ & $28.69$ \\ 
   \textit{Our Prompting-based} \\
        \quad \directsumm & Llama-2-7B  & $12.49$ & $32.63$ & $21.85$ & $45.32$ & $7.45$ \\ 
        \quad \directsumm & GPT-3.5  & $33.58$ & $57.02$ & $60.18$  & $53.93$ & $22.21$ \\
        \quad \reasonsumm & Llama-2-7B & $16.45$ & $37.13$ & $25.13$ & $48.16$ & $12.17$ \\ 
        \quad \reasonsumm & GPT-3.5  & $40.84$ & $62.68$ & $64.98$  & $56.24$ & $24.36$ \\
         \quad \reasonsumm & GPT-4    & $42.32$ & $64.36$ & $67.36$  & $\textbf{66.83}$ & $\textbf{32.37}$ \\
    \bottomrule
\end{tabular}
} 
\end{adjustbox}
\end{table*}

\begin{table}[t]
    \caption{Human evaluations of representative models on the test set. Three expert annotators are recruited to evaluate $100$ random examples for each model. The best results are in \textbf{bold}.}
    \label{tab:pred_human_eval}
    \centering
    \begin{adjustbox}{max width=1.0\textwidth} {
      \begin{tabular}{l *{3}{c}}
        \toprule
        \textbf{Model} & \textbf{Backbone}  & \textbf{Faithfulness}   & \textbf{Fluency} \\
        \midrule
        \omnitablargesum & BART-large  & 0.19 & 4.79 \\
        \reasonsumm & GPT-3.5       & 0.28 & 4.72 \\
        \reasonsumm   & GPT-4    & \textbf{0.56} & \textbf{4.84} \\
        \bottomrule
        \end{tabular}
    } 
    \end{adjustbox}
\end{table}
\begin{table}[t]
    \caption{Comparisons between single-table and multi-table examples on the test set. R-L, BSc, and PA stand for ROUGE-L, BERTScore, and PARENT, respectively.}
    \label{tab:num_table_results}
    \centering
    \small
    \begin{adjustbox}{max width=\columnwidth} {
    \begin{tabular}{l ccc ccc}
    \toprule
    \multirow{3}{*}{\textbf{Model}}    & \multicolumn{6}{c}{\textbf{Number of Tables}} \\
    \cmidrule{2-7}
            & \multicolumn{3}{c}{\textbf{Single-Table}} & \multicolumn{3}{c}{\textbf{Multi-Table}} \\
    \cmidrule(r){2-4}\cmidrule{5-7}
      &  \textbf{R-L} & \textbf{BSc} & \textbf{PA} & \textbf{R-L} & \textbf{BSc} & \textbf{PA} \\
    \midrule
    
    \omnitablargesum (BART-large)    & $70.73$  & $72.11$  & $37.73$ & $65.67$ & $67.20$ & $21.56$ \\
    \reasonsumm (GPT-3.5)  & $64.95$  & $67.04$ & $28.92$  & $61.61$ & $64.18$ & $19.29$ \\
    \reasonsumm (GPT-4)    & $66.03$  & $69.23$ & $38.11$  & $62.58$ & $64.98$ & $29.87$ \\
    \bottomrule
    \end{tabular}
   } 
    \end{adjustbox}
\end{table}

\subsection{\ours Evaluation}
\label{subsec:metrics}

To assess model performance, we employ automated and human evaluations. For automated evaluation, we measure the quality of the generated summary based on the corresponding reference summary and the execution table, as detailed in Subsection \ref{subsec:annotation}. For human evaluation, we evaluate two key aspects: \textit{faithfulness} and \textit{fluency}.

\paragraph{Text-Based Automated Evaluation} 

We first evaluate the quality of a generated summary w.r.t.\ the corresponding reference textual summary by estimating the similarity between them in general aspects, such as fluency and accuracy.
Following~\citet{zhao-etal-2023-qtsumm}, we adopt two lexical-based metrics, SacreBLEU~\cite{papineni-etal-2002-bleu} and ROUGE-L (longest common sub-sequences)~\cite{lin-hovy-2003-automatic}, along with a semantic-based metric, BERTScore~\cite{zhang2020bertscore}. 
We report the F1 versions for both ROUGE-L and BERTScore. We use \texttt{deberta-xlarge-mnli}~\cite{he2021deberta} as the backbone for BERTScore.

\paragraph{Table-Based Automated Evaluation} 

In contrast to text-based evaluations, table-based evaluations focus more on specific aspects, such as completeness and faithfulness of the generated summary. To assess these aspects, we employ two metrics: String Exact Match (STR-EM) \cite{stelmakh-etal-2022-asqa} and PARENT \cite{dhingra-etal-2019-handling}. STR-EM quantifies the proportion of facts or entities from the execution table that are accurately represented in the generated summary. PARENT evaluates summary completeness by integrating both the reference summary and the execution table. PARENT has demonstrated a significant correlation with human judgments.

\paragraph{Human Evaluation} 

In addition to the automated evaluations, we conduct human evaluations, specifically targeting \textit{faithfulness} and \textit{fluency}. These evaluations follow the detailed annotation guidelines specified in Subsection \ref{subsec:quality_eval}. For each model, $100$ generated summaries are randomly sampled from the test set, and their quality was assessed by three expert annotators.

\section{Results and Analysis}
\label{sec:results}

\paragraph{Main Results}

We present the summarization performance of various models in \autoref{tab:main_results}. 
We observe that our \reasonsumm methods markedly outperform the \directsumm approach in both text-based and table-based evaluations. For instance, when we use GPT-3.5 as the backbone, \reasonsumm surpasses \directsumm by about $5$ points regarding BERTScore ($64.98$ vs. $60.18$).  This is because the \reasonsumm method tackles the task into sequential phases of table reasoning and summarization. This specialized focus on multi-table reasoning in the initial phase enables the generation of more query-relevant facts and entities, thus enhancing the LLM's reasoning ability. Consequently, the subsequent summarization task benefits from generating more relevant facts, yielding superior performance compared to the \directsumm approach, which jointly addresses table reasoning and summarization tasks.
Additionally, our results indicate that baseline models fine-tuned on our training dataset largely surpass the \directsumm methods employing GPT-3.5 or Llama-2 as backbones in text-based evaluations. In comparison, our \reasonsumm approaches using GPT-3.5 and GPT-4 as backbones demonstrate competitive performance alongside the \bartlargesum and \multitabqasum. This also indicates the effectiveness of \reasonsumm. Furthermore, the \llamasum model achieves the highest performance. Particularly in text-based evaluation, when compared to the \reasonsumm with the backbone Llama-2, the fine-tuned \llamasum shows significant improvements, underscoring our dataset's efficacy as a robust training resource for query-focused multi-table summarization scenarios.
Conversely, in table-based evaluations, however, the trend reverses. For instance, our \reasonsumm method employing GPT-4 as the backbone, largely outperforms the leading \llamasum model. This discrepancy indicates that while smaller, fine-tuned models may produce plausible summaries, they lack proficiency in table reasoning across multiple tables to gather query-relevant facts, thereby leading to inferior performance on table-based metrics. In comparison, our \reasonsumm methods produce more facts relevant to the queries, demonstrating superior performance in terms of table reasoning ability.

\paragraph{Human Evaluation}

\autoref{tab:pred_human_eval} illustrates the results of sampled human evaluation on the test set. Despite lower scores on automated text-based evaluation metrics such as SacreBLEU and ROUGE scores, our \reasonsumm method, which integrates GPT-4, significantly outperforms the fine-tuned \omnitablargesum in human evaluations, particularly in terms of faithfulness ($0.56$ vs. $0.19$). This disparity highlights the superior reasoning ability of our methods over fine-tuned baseline models.
Furthermore, our findings reveal a mismatch between text-based automated metrics and human evaluations, aligning with the observations made by \citet{zhao-etal-2023-qtsumm}. In contrast, our table-based metrics demonstrate a strong correlation with human judgments concerning faithfulness. This indicates that table-based evaluations are complementary to text-based evaluations, enabling a more comprehensive evaluation for system performance comparison.

\paragraph{Single- vs. Multi-Table}

To enhance our understanding of the challenges presented in multi-table scenarios, we conducted a performance comparison between single-table and multi-table inputs from our test set, as illustrated in \autoref{tab:num_table_results}. It is worth noting that approximately $30\%$ of the examples in our dataset are characterized by single-table inputs. Our analysis reveals that the presence of multiple input tables significantly deteriorates the performance across all evaluated metrics for every model tested. This decline in performance is particularly significant for the smaller \omnitablargesum, whereas it is least noticeable for our method utilizing GPT-4 as the backbone.
For instance, considering PARENT scores, the decrease observed with \omnitablargesum is approximately $16$ points, moving from $37.73$ to $21.56$. In contrast, our \reasonsumm shows a more modest decrease of about $8$ points, dropping from $38.11$ to $29.87$. This reduction is nearly half that observed with \omnitablargesum.
These findings suggest that while multi-table reasoning poses greater challenges compared to single-table scenarios, increased model capacity can effectively narrow this performance gap.

\paragraph{Qualitative Analysis}
\label{subsec:quali_analysis}

To enhance our understanding of the strengths of our approach and challenges within the task, we conduct a manual analysis of the summaries generated by \reasonsumm with the backbone of GPT-3.5 on the test set, including success and failure cases. We include more case studies in Appendix~\ref{app_sec:case_study}.  We observe that our method successfully performs arithmetic and multi-table operations in some cases.
A success case illustrates the strengths of our approach. For the query ``\textit{Which employee received the most awards in evaluations? Give me the employee name.}'' over two input tables:

\begin{center}
\begin{minipage}[c]{0.5\columnwidth}
\centering
\small
\captionsetup{labelformat=empty}
 \captionof{table}{\textbf{Employee}}
\begin{tabular}{|c|c|c|c|}
\hline
ID & Name & Age  \\
\hline
\textbf{1}  & \textbf{\makecell{George\\ Chuter}} & 23 \\
2  & \makecell{Lee \\Mears}  & 29  \\ 
... & ... & ... \\ 
\hline 
\end{tabular}
\end{minipage}
\begin{minipage}[c]{0.48\columnwidth}
\small
\captionsetup{labelformat=empty}
 \captionof{table}{\textbf{Evaluation}}
\centering
\begin{tabular}{|c|c|c|}
\hline
 ID & \makecell{Year\\\_awarded}  & Bonus \\
 \hline
\textbf{1} & 2011 & \textbf{3000}  \\ 
2 & 2015 & 3200  \\
\textbf{1} & 2016 & \textbf{2900}  \\
... & ... & ...  \\
\hline 
\end{tabular}
\end{minipage}
\end{center}

\smallskip\noindent%
With the reference summary ``\textit{The recipient of the most awards in evaluations is \textcolor{blue}{George Chuter}.}'', our method reasons over the $2$ tables, performing complex table operations, such as \textit{count} and \textit{join}. Specifically, the method finds two records of awards of \textcolor{blue}{George Chuter} in the table \texttt{Evaluation} and aggregates the total number of awards. After joining the two tables, the method accurately identifies \textcolor{blue}{George Chuter} as the person with the most awards, generating ``\textit{The employee who received the most awards in evaluations is \textcolor{blue}{George Chuter}.}''

A failure case illustrates the challenges of multi-table scenarios. Consider the query ``\textit{What are the names of all European countries with at least 3 manufacturers?}'' over three input tables:

\begin{center}
\begin{minipage}[c]{0.4\columnwidth}
\centering
\small
\captionsetup{labelformat=empty}
\captionof{table}{\textbf{Continents}}
\begin{tabular}{|c|c|}
\hline
\makecell{Cont\\Id} & Continent  \\
\hline
 1 & America \\
 \textbf{2} & \textbf{Europe}  \\
 3 & Asia  \\ 
 4 & Africa  \\
 5 & Australia \\
\hline 
\end{tabular}
\end{minipage}
\begin{minipage}[c]{0.5\columnwidth}
\small
\captionsetup{labelformat=empty}
 \captionof{table}{\textbf{Countries}}
\centering
\begin{tabular}{|c|c|c|}
\hline
\makecell{Country\\Id} & \makecell{Country\\Name} & \makecell{Cont-\\inent}\\
\hline
 \textbf{2} & \textbf{Germany} & \textbf{2}  \\
 \textbf{3} & \textbf{France} & \textbf{2}  \\
1 & {USA} & 1 \\
\textbf{8} & \textbf{Korea} & \textbf{3}  \\
... & ... & ... \\
\hline 
\end{tabular}
\end{minipage}
\end{center}

\begin{center}
\begin{minipage}[c]{1\columnwidth}
\small
\captionsetup{labelformat=empty}
 \captionof{table}{\textbf{Car Makers}}
\centering
\begin{tabular}{|c|c|c|c|}
\hline
Id & Maker & Full Name & Country\\
\hline
 2 & Volkswagen & Volkswagen & \textbf{2}  \\ 
3 & bmw & BMW & \textbf{2}  \\ 
... & ... & ... & ... \\
 7 & citroen & Citroen & \textbf{3}  \\
... & ... & ... & ... \\
 14 & opel & Opel & \textbf{2}  \\ 
 15 & peugeaut & Peugeaut & \textbf{3}  \\ 
 16 & renault & Renault & \textbf{3}  \\
... & ... & ... & ...\\
 22 & kia & Kia Motors & \textbf{8}  \\ 
\hline 
\end{tabular}
\end{minipage}
\end{center}

\smallskip\noindent%
With the reference summary ``\textit{There are 2 European countries with at least 3 manufacturers. The names of these countries are \textcolor{blue}{France} and \textcolor{blue}{Germany}.}'', in which the correct country names are marked in \textcolor{blue}{blue}. The method incorrectly generates ``\textit{There are 2 European countries which have at least 3 manufacturers. Their names are \textcolor{blue}{France} and \textcolor{red}{Korea}.}'', in which the incorrect country name is marked in \textcolor{red}{red}. Even though this generated summary exhibits a high degree of \textit{fluency}, it is only partially \textit{faithful} and \textit{complete} due to the incorrect inclusion of \textit{\textcolor{red}{Korea}}, a country which is not located in Europe. 
This case exemplifies the complexity and challenges of multi-table operations since the proposed approach struggles to combine information from the three tables based on the corresponding column headers, \texttt{Country ID} and \texttt{Cont Id}.

\section{Conclusion}
\label{sec:conclu}

In conclusion, this work addresses the shortcomings of current table summarization techniques through the introduction of an innovative method for query-focused multi-table summarization. Our proposed method leverages user queries and analyzes multiple tables to generate summaries that directly cater to users' information needs. Additionally, we make a significant contribution to the field by presenting a comprehensive dataset tailored explicitly for this query-focused multi-table summarization task, thereby enabling further research. Extensive evaluations conducted demonstrate the superior performance of our method compared to existing baselines, underscoring the complexities associated with accurate summarization in the context of intricate table reasoning. Overall, our work not only propels advancements in query-focused multi-table summarization but also offers valuable insights to guide future exploration and development in this field.

\section*{Limitations}

Our proposed method, \reasonsumm, demonstrates competitive performance relative to fine-tuned models. However, it could be further enhanced by addressing specific multi-table reasoning operations, such as \textit{join}, \textit{union}, and \textit{intersect}. Future versions of our method could benefit from the development of specialized prompting strategies tailored to each of these operations, which is a promising direction for future research.
Our evaluation reveals potential room for improvement. In our experiments, we observe a significant divergence between text-based metrics and table-based metrics.
This divergence suggests that current text-based metrics may not adequately capture the accuracy of generated summaries. 
Enhancements in these metrics are necessary to ensure more reliable evaluations. 
Additionally, the development of unified evaluations that integrate both text-based and table-based evaluations could guide future research directions.

\begin{ack}
This research was supported by the China Scholarship Council
(CSC) under grant number 202008440470. 
The views expressed in the content belong solely to the authors and may not reflect the perspectives or endorsements of their respective employers or sponsors.
\end{ack}

\bibliography{main}

\appendix

\section{Details of Dataset Construction}

\subsection{Details of Human Post-Correction}
\label{app_subsec:correction_details}

The human post-correction process consists of two pivotal steps.
In the initial phase, we ensure that all summaries in both the validation and test sets adhere strictly to the criteria of completeness and faithfulness relative to the corresponding execution output tables, based on Subsection \ref{subsec:quality_eval}. 
During the annotation process, it has been observed that certain summaries are either incomplete or contain factual discrepancies. Notably, some instances exhibit both deficiencies. Consequently, we meticulously review all annotations to rectify any summaries that are either incomplete or unfaithful.

\noindent
Consider the following execution output table, which consists of five rows and two columns: \\

\begin{center}
\begin{minipage}[c]{\columnwidth}
\small
\centering
\begin{tabular}{|c|c|}
\hline
Name & Location \\
\hline
Auto Club Speedway & Fontana, CA \\
Daytona International Speedway & Daytona Beach, FL \\
Homestead-Miami Speedway & Homestead, FL \\
Kansas Speedway & Kansas City, KS \\
Martinsville Speedway & Ridgeway, VA \\
\hline 
\end{tabular}
\end{minipage} 
\end{center}

\noindent \\
An incomplete annotated summary states: ``\textit{There are 4 tracks that have had exactly 1 race. The names and locations of these tracks are Auto Club Speedway in Fontana, CA, Daytona International Speedway in Daytona Beach, FL, Homestead-Miami Speedway in Homestead, FL, and Kansas Speedway in Kansas City, KS.}'' summary erroneously omits the last track \textit{Martinsville Speedway, Ridgeway, VA}.
An unfaithful summary states: ``\textit{There are 5 tracks that have had exactly 1 race. The names and locations of these tracks are Auto Club Speedway in Fontana, CA, Daytona International Speedway in Daytona Beach, FL, Homestead-Miami Speedway in Homestead, FL, Kansas Speedway in Kansas City, KS, and Martinsville Speedway in Kansas City, KS.}'' 
This summary inaccurately locates Martinsville Speedway in \textit{Kansas City, KS} instead of its actual location in \textit{Ridgeway, VA}.

Subsequently, to mitigate bias in the annotation process, we address potential biases stemming from using outputs generated by our annotators \annollm to create validation and test sets.
This is critical when the same LLM is used as a baseline model, which may inflate performance metrics.
Our mitigation strategy involves manually revising each summary to enhance readability and vary vocabulary without altering factual content, such as proper nouns and numbers. 
For instance, the initial summary ``\textit{There are 5 high schoolers who have a grade of over 5 and have 2 or more friends. Their names are Alexis, Kris, Jordan, Gabriel, and Andrew.}'' is rephrased to ``\textit{5 high school students have a grade above 5 and at least 2 friends. They are Alexis, Kris, Jordan, Gabriel, and Andrew.}''

\section{Case Study}
\label{app_sec:case_study}

We conducted case studies by randomly selecting generated summaries from our \reasonsumm (GPT-4) model on the test set. The results are presented in \autoref{tab:case_study}. Compared to the gold summaries, while the generated summaries are generally longer, they frequently suffer from hallucinations, where incorrect information is introduced. For example, in the first case, the gold summary states: ``Five high school students have a grade above 5 and at least two friends. They are Alexis, Kris, Jordan, Gabriel, and Andrew.'' However, the model-generated summary incorrectly reports: ``There are 15 high schoolers who have a grade of over 5 and have 2 or more friends. These students are Jordan, Gabriel, Tiffany, Cassandra, Andrew, Kris, Brittany, Alexis, Austin, Gabriel, Jessica, Jordan, John, Kyle, and Logan.'' This summary introduces incorrect names, such as Logan,  which are not part of the correct answer to the given query.

\begin{table*}[h]
\centering
\begin{ADLinactivate}
\resizebox{\columnwidth}{!}{%
\footnotesize
\begin{tabular}{p{3cm}|p{4cm}|p{4cm}|p{4cm}}
\toprule
 Query & Execution Table & Gold Summary & \reasonsumm (GPT-4) \\
\midrule
\textit{Q}: What are the names of high schoolers who have a grade of over 5 and have 2 or more friends? 
&
\texttt{col : name row 1 : Alexis row 2 : Kris row 3 : Jordan row 4 : Gabriel row 5 : Andrew} 
&
Five high school students have a grade above 5 and at least two friends. They are Alexis, Kris, Jordan, Gabriel, and Andrew.
& 
There are 15 high schoolers who have a grade of over 5 and have 2 or more friends. These students are Jordan, Gabriel, Tiffany, Cassandra, Andrew, Kris, Brittany, Alexis, Austin, Gabriel, Jessica, Jordan, John, Kyle, and Logan. \\ 
\midrule

\textit{Q}: Show names of teachers and the courses they are arranged to teach.
&
\texttt{col : Name | Course row 1 : Kearsley Brown | Math row 2 : Vicente Carretero | Math row 3 : Gustaaf Deloor | Science row 4 : Anne Walker | History row 5 : Anne Walker | Bible row 6 : Lucy Wong | Musi}
&
The school has a total of 6 teachers, each assigned to specific courses. Kearsley Brown and Vicente Carretero are responsible for teaching Math. Gustaaf Deloor handles the Science classes. Anne Walker teaches both History and Bible. Finally, Lucy Wong is in charge of the Music classes.
 & 
There are five teachers in total and each is arranged to teach a specific course. Kearsley Brown and Vicente Carretero are both arranged to teach Math. Gustaaf Deloor is arranged to teach Science. Anne Walker is arranged to teach History and Bible. Lastly, Lucy Wong is arranged to teach Music.
\\ 
 \midrule
 \textit{Q}: What are the names of poker players in descending order of earnings?
&
\texttt{col : Name row 1 : Maksim Botin row 2 : Aleksey Ostapenko row 3 : Teodor Salparov row 4 : Semen Poltavskiy row 5 : Yevgeni Sivozhelez}
&
There are a total of 5 poker players. Listed in descending order based on their earnings, the players are Maksim Botin, Aleksey Ostapenko, Teodor Salparov, Semen Poltavskiy, and Yevgeni Sivozhelez.
& 
There are 5 poker players listed in descending order of earnings. These players are Maksim Botin, Aleksey Ostapenko, Semen Poltavskiy, Teodor Salparov, and Yevgeni Sivozhelez.
 \\ 
 \midrule
\textit{Q}: For each stadium, how many concerts play there?
&
\texttt{col : Name | count(*) row 1 : Stark's Park | 1 row 2 : Glebe Park | 1 row 3 : Somerset Park | 2 row 4 : Recreation Park | 1 row 5 : Balmoor | 1}
&
There are 5 stadiums in total, hosting varying numbers of concerts. Stark's Park, Glebe Park, Recreation Park, and Balmoor host 1 concert. Somerset Park hosts 2 concerts.
& 
The number of concerts played at each stadium varies. Stark's Park hosts 1 concert, Somerset Park hosts 2 concerts, while Bayview Stadium, Hampden Park, Forthbank Stadium, and Gayfield Park do not host any concerts. Recreation Park, Balmoor, and Glebe Park each host 1 concert.
 \\ 
\bottomrule
\end{tabular}%
}
\end{ADLinactivate}
\caption{Case studies on the test set with generated summaries by our \reasonsumm (GPT-4). We omit input tables as they may be too large to display.}
\label{tab:case_study}
\end{table*}

\section{Detailed Prompts}
\label{app_sec:propmt}

To facilitate reproducibility, we have included all detailed prompts for our proposed methods. The prompt corresponding to \directsumm is illustrated in Figure~\ref{prompt:direct-summ-long}, while the prompts for \reasonsumm are provided in Figures \ref{prompt:multi-stage-1} and \ref{prompt:multi-stage-2}.

\begin{figure*}[h]
\begin{prompt}[title={Figure \thetcbcounter: Direction Summarization}, label=prompt:direct-summ-long]
    \textbf{Instruction:} You will be given a query along with one or more tables to complete two tasks step by step. Each table contains a name and content with multiple rows and columns, formatted as follows: \\
    col: <column header 1> | <column header 2> | ... | <column header n> row 1: <value 1,1> | <value 1,2> | ... | <value 1,n> row 2: <value 2,1> | <value 2,2> | ... | <value 2,n> ... row m: <value m,1> | <value m,2> | ... | <value m,n>. \\
    \\
    \textbf{Task 1: Reasoning over Tables.} \\
    Your first task is to perform table-based reasoning to obtain the query-relevant facts across the tables, such as numerical data and entities. This may involve performing arithmetic calculations and combining data from multiple tables if necessary. Please begin your response with "Facts:" and enumerate all discovered facts one by one, separating them with commas ",". Let's think step by step. \\
    \\
    \textbf{Task 2: Generating a Summary.} \\
    Your second task is to write a concise, fluent, and accurate summary based on the query-relevant facts generated in the first task. This summary should begin with the word "Summary:" and follow the guidelines as follows: 1) Introduction: Begin by using a numeral to indicate the total number of the facts if there are two or more; Then, rephrase the query as a declarative statement while retaining all relevant keywords. 2) Body: Present all discovered facts one by one. The summary should be a standard paragraph format without using lists, containing a minimum of 5 words but not exceeding 300 words in length. \\
    You can refer to the demonstrations below. Each demonstration consists of a query, tables, and human-written facts, and a summary. \\
    \\
    \textbf{Demonstrations:} \\
    Query: Show the name for regions not affected. \\
    Table $1$: Name: region; Content: col : Region\_id | Region\_code | Region\_name row 1 : 1 | AF | Afghanistan row 2 : 2 | AL | Albania row 3 : 3 | DZ | Algeria row 4 : 4 | DS | American Samoa row 5 : 5 | AD | Andorra row 6 : 6 | AO | Angola row 7 : 7 | AI | Anguilla row 8 : 8 | AQ | Antarctica row 9 : 9 | AG | Antigua and Barbuda row 10 : 10 | CY | Cyprus row 11 : 11 | CZ | Czech Republic row 12 : 12 | DK | Denmark row 13 : 13 | DJ | Djibouti \\
    Table $2$: Name: Affected Region; Content: col : Region\_id | Storm\_ID | Number\_city\_affected row 1 : 1 | 1 | 10 row 2 : 2 | 1 | 15 row 3 : 3 | 3 | 30 row 4 : 1 | 4 | 22 row 5 : 12 | 5 | 37 row 6 : 2 | 5 | 12 \\ 
    Facts: American Samoa, Andorra, Angola, Anguilla, Antarctica, Antigua and Barbuda, Cyprus, Czech Republic, and Djibouti are the names for regions not affected. \\
    Summary: There are 9 regions that are not affected. These regions include American Samoa, Andorra, Angola, Anguilla, Antarctica, Antigua and Barbuda, Cyprus, Czech Republic, and Djibouti. \\
    \ldots  \ldots \\
    \\
    Now follow the instructions and the demonstrated style above to complete the two tasks step by step for the query and tables provided below: \\
    \\
    Query: \emph{input query here} \\
    Tables: \emph{input tables here} \\
\end{prompt}
\end{figure*}

\begin{figure*}[h]
\begin{prompt}[title={Figure \thetcbcounter: Reason-then-Summarize Phase 1}, label=prompt:multi-stage-1]
    \textbf{Instruction:} You will be given a query along with one or more tables to complete the task below. Each table contains a name and content with multiple rows and columns, formatted as follows: \\
    col: <column header 1> | <column header 2> | ... | <column header n> row 1: <value 1,1> | <value 1,2> | ... | <value 1,n> row 2: <value 2,1> | <value 2,2> | ... | <value 2,n> ... row m: <value m,1> | <value m,2> | ... | <value m,n>. \\
    \\
    \textbf{Task: Reasoning over Tables.} \\
    Your task is to perform table-based reasoning to obtain query-relevant facts using only the content from the input tables, such as numerical data and entities. based on the query. This may involve performing arithmetic calculations and combining data from multiple tables if necessary. Please begin your response with 'Facts:' and enumerate all discovered facts one by one, separating them with commas ','. Let's think step by step. \\   
    \\
    \textbf{Demonstrations:} \\
    Query: Show the name for regions not affected. \\
    Table $1$: Name: region; Content: col : Region\_id | Region\_code | Region\_name row 1 : 1 | AF | Afghanistan row 2 : 2 | AL | Albania row 3 : 3 | DZ | Algeria row 4 : 4 | DS | American Samoa row 5 : 5 | AD | Andorra row 6 : 6 | AO | Angola row 7 : 7 | AI | Anguilla row 8 : 8 | AQ | Antarctica row 9 : 9 | AG | Antigua and Barbuda row 10 : 10 | CY | Cyprus row 11 : 11 | CZ | Czech Republic row 12 : 12 | DK | Denmark row 13 : 13 | DJ | Djibouti \\
    Table $2$: Name: Affected Region; Content: col : Region\_id | Storm\_ID | Number\_city\_affected row 1 : 1 | 1 | 10 row 2 : 2 | 1 | 15 row 3 : 3 | 3 | 30 row 4 : 1 | 4 | 22 row 5 : 12 | 5 | 37 row 6 : 2 | 5 | 12 \\ 
    Facts: American Samoa, Andorra, Angola, Anguilla, Antarctica, Antigua and Barbuda, Cyprus, Czech Republic, and Djibouti are the names for regions not affected. \\
    \ldots  \ldots \\
    \\
    Now follow the instructions and the demonstrated style above to complete the task step by step for the query and tables provided below: \\
    \\
    Query: \emph{input query here} \\
    Tables: \emph{input tables here} \\
\end{prompt}
\begin{prompt}[title={Figure \thetcbcounter: Reason-then-Summarize Phase 2}, label=prompt:multi-stage-2]
    \textbf{Instruction:} You will be given a query along with the generated facts from phase 1 to complete the task below. \\
    \\
    \textbf{Task: Generating a Summary.} \\
    Your task is to write a concise, fluent, and accurate summary based on the query-relevant facts generated in the first task. This summary should begin with the word "Summary:" and follow the guidelines as follows: 1) Introduction: Begin by using a numeral to indicate the total number of facts if there are two or more; Then, rephrase the query as a declarative statement while retaining all relevant keywords. 2) Body: Present all discovered facts one by one. The summary should be a standard paragraph format without using lists, containing a minimum of 5 words but not exceeding 300 words in length. \\
    You can refer to the demonstrations below. Each demonstration consists of a query, tables, and human-written facts, and a summary. \\
    \\
    \textbf{Demonstrations:} \\
    Query: Show the name for regions not affected. \\
    Facts: American Samoa, Andorra, Angola, Anguilla, Antarctica, Antigua and Barbuda, Cyprus, Czech Republic, and Djibouti are the names of regions not affected. \\
    Summary: There are 9 regions that are not affected. These regions include American Samoa, Andorra, Angola, Anguilla, Antarctica, Antigua and Barbuda, Cyprus, Czech Republic, and Djibouti. \\
    \ldots  \ldots \\
    \\
    Now follow the instructions and the demonstrated style above to complete the task step by step for the query and generated facts from phase 1 provided below: \\
    \\
    Query: \emph{input query here} \\
    Table: \emph{generated facts here} \\
\end{prompt}
\end{figure*}

\end{document}